\documentclass{article}
\usepackage{spconf,amsmath,graphicx}
\usepackage{diagbox}
\usepackage{amssymb}
\usepackage{bbding}
\usepackage{multirow}
\usepackage{float}
\usepackage{longtable}
\usepackage[table,xcdraw]{xcolor}
\usepackage{multirow}
\usepackage{support-caption}
\usepackage[skip=0pt]{caption}
\usepackage{subcaption}
\usepackage{graphicx}

\title{Relationship-based Neural Baby Talk}
 \name{
 Fan Fu$^{1 }$ 
 \qquad Tingting Xie$^{2}$  
 \qquad Ioannis Patras$^{2}$ 
 \qquad Sepehr Jalali$^{1,3}$
 }
 \address{$^{1 }$ City, University of London \qquad
 $^{2}$ Queen Mary University of London \qquad
 $^{3}$ MediaTek Research \qquad}

\begin{document}
%
\maketitle
%


\begin{abstract}


Understanding interactions between objects in an image is an important element for generating captions. In this paper, we propose a relationship-based neural baby talk (R-NBT) model to comprehensively investigate several types of pairwise object interactions by encoding each image via three different relationship-based graph attention networks (GATs). We study three main relationships: \textit{spatial relationships} to explore geometric interactions, \textit{semantic relationships} to extract semantic interactions, and \textit{implicit relationships} to capture hidden information that could not be modelled explicitly as above. We construct three relationship graphs with the objects in an image as nodes, and the mutual relationships of pairwise objects as edges. By exploring features of neighbouring regions individually via GATs, we integrate different types of relationships into visual features of each node. Experiments on COCO dataset show that our proposed R-NBT model outperforms state-of-the-art models trained on COCO dataset in three image caption generation tasks.

\end{abstract}
\section{Introduction}
\label{sec:intro}

Image caption generation is a challenging task in the intersection of computer vision and natural language processing. It aims at generating a natural readable sentence (caption) to summarise a given image. This has various real-world applications such as caption-based image retrieval, smart wearable devices, and human-computer interaction.


Similar to machine translation, image caption generation methods usually leverage an \textit{Encoder-Decoder framework} to translate an image into text~\cite{vinyals2015show,xu2015show,he2019vd}. The \textit{Encoder}, typically a CNN~\cite{szegedy2015going,he2016deep}, is utilised to obtain image representations; the \textit{Decoder}, built up from RNN~\cite{hochreiter1997long}, is used to generate sentences dynamically. \cite{xu2015show,lu2017knowing} further introduce an attention mechanism in the decoder to automatically identify salient image regions corresponding to each emitted word. Rather than compressing an entire image into a static representation, \cite{lu2018neural,anderson2018bottom,zohourianshahzadi2020neural} propose to extract a set of salient image regions from Faster R-CNN~\cite{ren2015faster} and feed them to the decoder.

However, an optimal caption generation network should capture not only what objects are present in an image, but also the relationships between them. GCN-LSTM~\cite{yao2018exploring} is the first to propose integration of two types of relationships -- spatial and semantic relationships -- into the encoder via a Graph Convolutional Network (GCN)~\cite{kipf2016semi}. It constructs two separate graphs for two relationships, each takes a set of image regions as nodes, and the relationships as edges. This allows to enhance the region-level features by attending over spatial-/semantic-related nodes via the GCN. 

The main limitation of GCN-LSTM is that semantic and spatial relationships cannot model all types of object interactions, such as object counting. Inspired by~\cite{hu2018relation,li2019relation}, we propose adding an extra implicit relationship graph to model the implicit relationships which cannot be captured by semantic and spatial relationships explicitly. Particularly, based on Neural Baby Talk (NBT)~\cite{lu2018neural}, we present our relationship-based Neural Baby Talk (R-NBT) model combining all visual relationships via three graph structures. Instead of using GCNs as in~\cite{yao2018exploring}, which treats all neighbour nodes equally, we exploit graph attention networks (GATs) to assign different weights to different nodes to focus more on closer neighbours.

In this paper we present three main contributions: 
\\
1- We integrate implicit relationship learning into the NBT model in addition to the explicit relationship encoding (i.e., spatial and semantic relationships). 
\\
2-  We implement this by embedding graph attention networks in image caption generation tasks for the first time and utilise graph attention mechanism to enable better region-level feature representations by observing entire neighbourhood differently to study the mutual impact between pairwise nodes.
\\
3- We implement our proposed model on three different image captioning tasks on COCO dataset and demonstrate new state of the art results on models trained on COCO dataset.

\begin{figure*}
    \centering
    \begin{center}
        \includegraphics[width=1.1\textwidth]{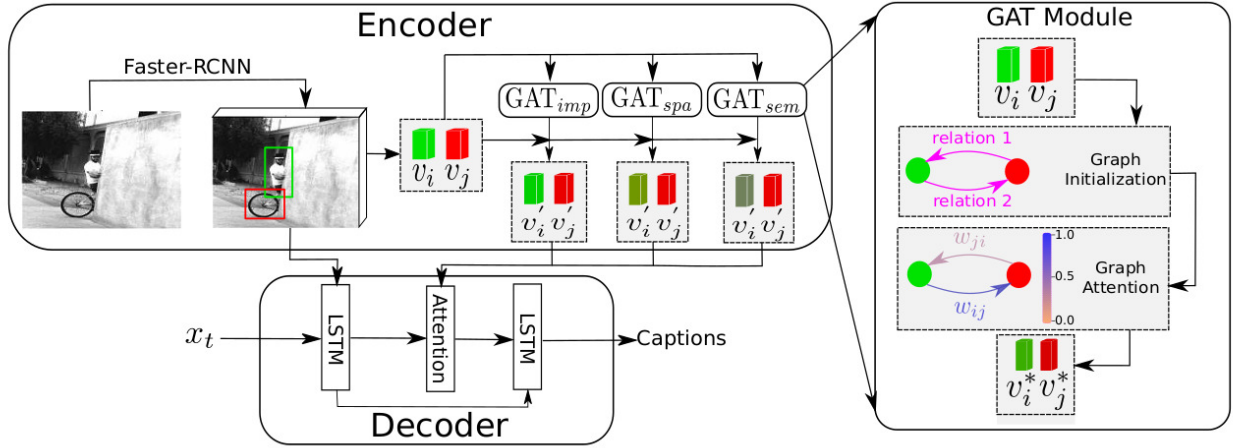}
    \end{center}
    \vspace{-1.8cm}
    \caption{An overview of our R-NBT. Faster R-CNN is used to obtain region-level features of objects, then feed them into graph attention networks to integrate different relationships into each region representation, by attending over neighbouring region-level features (i.e. connected nodes with the corresponding relationship). The original features and relationship features are then summed up, together with the entire image features and last predicted word, are fed into the decoder to generate sentences.}
    \label{fig:overview}
\end{figure*}

\section{Method}
\label{sec:method}

\textbf{Baseline.} In this paper, we propose the Relationship-based Neural Baby Talk (R-NBT) model, built on a state-of-the-art method, Neural Baby Talk (NBT) \cite{lu2018neural}, as shown in Fig.~\ref{fig:overview}. It adopts the typical Encoder-Decoder framework, by taking Faster R-CNN~\cite{ren2015faster} pre-trained with ResNet-101~\cite{he2016deep} as encoder to produce the region-level features, and two LSTM layers with attention mechanism as decoder to generate the captions. Unlike NBT, we further enhance the region-level features by feeding into three different relationship-based graph attention networks (GATs), while spatial/semantic GATs are constructed with prior knowledge of spatial/semantic relationships and implicit GAT is used for modelling other relationships. The whole network is mainly optimised by a cross entropy loss. In this paper, we mainly focus on the encoder part. In the following, we first formulate the problem, explaining both the graph construction and the relationship extractions. Then we introduce graph attention networks, and finally describe the fusion strategy to integrate different relationship-based GATs.

\subsection{Problem Formulation}

As shown in Fig.~\ref{fig:overview}, given an image $I$, we pass it through Faster R-CNN pre-trained with ResNet-101 to obtain a set of detected objects: $\mathbf{O}=\{\mathbf{o}_i\}^n_{i=1}$ and corresponding region-level features of objects: $V=\{\mathbf{v}_i\}^n_{i=1}$, where $n$ is the number of objects, $\mathbf{o}_i$ denotes the $i^{th}$ object, represented by a 5-dimensional vector of boundary position and object category, and $\mathbf{v}_i \in \mathbb{R}^{d}$ where $d$ is the dimension of features. By treating $V$ as nodes, we construct a relationship graph for each type of relationship, denoted as $G = (V, E)$, where $G \in \{G_{imp}, G_{spa}, G_{sem}\}$, $E \in \{E_{imp}, E_{spa}, E_{sem}\}$.

\textbf{Graph construction.} $G_{imp}$ is a fully-connected undirected graph with a set of $n(n-1)$ edges connecting every two nodes together, whose relationships (edges) are learned via graph attention mechanism implicitly. Different with $G_{imp}$, $G_{spa}$ and $G_{sem}$ are constructed with prior knowledge by assigning edges with spatial/semantic relationships. In particular, $G_{spa}$ is symmetrical and directional, for instance, if an edge from region $i$ to $j$ exists, the other one from object $j$ to $i$ will also exist, but with a different label. $G_{sem}$ is asymmetrical and directional, for instance: ``wearing" is a semantic relationship, since ``man wearing tie" is valid but there is no semantic relationship from tie to man.


\textbf{Relationship Extraction.} Usually a visual relationship is defined as a tuple $<object-predicate-object>$~\cite{lu2016visual}, where a relationship is equal to a predicate in this tuple. Following~\cite{yao2018exploring}, we classify spatial relationships into 11 classes and a no-relation class according to geometrical position, IoU, relative distance and angle of two objects. Regarding semantic relationships, we treat the semantic relationship extraction as a classification task and construct a classifier to predict it between any pairwise regions $i$ and $j$~\cite{yao2018exploring}. The classifier takes as input the region-level features of object $i$, $j$ and their corresponding union bounding box from Faster R-CNN, and outputs probabilities over pre-defined 16 semantic relationships (including no-relation) after two transformer layers. The trained classifier is then applied to predict semantic relationships for objects in images.



\textbf{Region-level Features Refinement.} By using each graph for image encoding to contextually refine the representations of an image region $v_i$ into relationship features $v^*_i$, we obtain the final refined region-level features of $i^{th}$ region $v^{'}_{i}$ by $v^{'}_i = v_i + v^{*}_i$. The $v^*_i$ will be introduced in Sect.~\ref{subsec:gan} in detail.

\subsection{Graph Attention Networks (GATs)}
\label{subsec:gan}

\textbf{Implicit GAT.} By modelling implicit relationships in GAT$_{imp}$, implicit relationship features $v^*_i$ are computed as a weighted sum of the region-level features at all connected regions. For each node, the relationship features computed by

\vspace*{-2mm}
\begin{equation}
\label{eq:relation}
    v^{*}_i = \sum_{1 \leq j \leq n} w_{ij} \cdot W v_j,     
\end{equation}

\noindent where $W \in \mathbb{R}^{d \times d}$ is the weights matrix of linear transformation, $w_{ij}$ is an attention coefficient to measure the impact of relationship from object $j$ to object $i$, which is computed as

\vspace*{-2mm}
\begin{equation}
\label{eq:attweight}
    w_{ij} = \frac {w^{b}_{ij} \cdot exp(w^v_{ij})}{\sum ^n_{k=1} w^b_{ik} \cdot exp(w^v_{ik})}.
\end{equation}

$w^v_{ij}$ is used to measure the similarity between object $i$ and object $j$, derived by $w^v_{ij} = (W_K \cdot v_i)^T \cdot W_Q \cdot v_j$, where $W_K$, $W_Q$ $\in \mathbb{R}^{d \times d}$ are transformation matrices. The aim of $w^b_{ij}$ is to measure the geometric position correlation between object $i$ and object $j$, computed by $w^b_{ij} = max \{0, W^b_G \cdot f^b_G(o_i, o_j)\}$.

Particularly, to compute $w^b_{ij}$ between two objects $i$ and $j$ with position ($x_i$, $y_i$, $w_i$, $h_i$) and ($x_j$, $y_j$, $w_j$, $h_j$), we firstly obtain 4-d relative geometry feature by
$$
f^b_G((log(\frac {|x_i - x_j|}{w_i}), log(\frac {|y_i - y_j|}{h_i}), log(\frac {w_j}{w_i}), log(\frac {h_j}{h_i}))^ T),
$$
and then project the 4-d features into high-dimensional representations by computing sine and cosine of different wavelengths~\cite{vaswani2017attention}. We then transform the high-dimensional representations (by a linear layer) into a scalar weight and trimmed at 0 for non-linearity projection.

\textbf{Spatial/Semantic GAT.} Different with implicit GAT, spatial/semantic relationship graphs are directional with a certain label. For making our model sensitive to both directionality and labels, we use separate transformation matrices and bias vectors for different directions and labels of edges, respectively, which following:

\vspace*{-5mm}
\begin{equation}
\label{eq:spatialfeature}
    v^{*}_i = \sum_{1 \leq j \leq n} w_{ij} \cdot (W_{dir_{(v_i,v_j)}}v_j + b_{lab_{(v_i,v_j)}}),
\end{equation}
\vspace*{-4mm}


\noindent where $b$ is a bias vector computed from the relationship graph and $W$ is a transformation matrix. $dir_{(v_i,v_j)}$ is used to select the transformation matrix with regard to the direction of each edge, such as $W_1$ for $v_i$-to-$v_j$, $W_2$ for $v_j$-to-$v_i$, and $W_3$ for $v_i$-to-$v_i$. The $lab_{(v_i, v_j)}$ indicates the class of edge connecting object $i$ to object $j$, and $w_{ij}$ is calculated as: 

\vspace*{-2mm}
\begin{equation}
    w_{ij} = \frac {exp(w^v_{ij})}{\sum ^K_{k=1} exp(w^v_{ik})},    
\end{equation}


\vspace*{-4mm}
\begin{equation}
\label{eq:attspatialweight}
w^v_{ij} = (W_Kv_i)^T \cdot W^v_{dir_{(v_i,v_j)}}v_j + c_{lab_{(v_i,v_j)}}
\end{equation}

\noindent where $W_K$ and $W^v_{dir_{(v_i,v_j)}}$ are transformation matrix.




\subsection{Multi-modal Fusion}

At the inference time, we adopt a weighted sum of predicted words at each time step to connect the three relationship GATs in our models. The finalised probability of each word at time step $t$ is computed as:
\begin{eqnarray}
\label{eq:fusion}
P_t = \alpha P^{spa}_t + \beta P^{sem}_t + (1 - \alpha - \beta) P^{imp}_t
\end{eqnarray}

where $P^{imp}_t$, $P^{spa}_t$ and $P^{sem}_t$ are the output probabilities of each word at time step $t$ from implicit, spatial and semantic GATs respectively, and $\alpha$ and $\beta$ are set as $0.3$ empirically. 


\section{Experiments}
\label{sec:exp}


To demonstrate the effectiveness of our R-NBT, we conduct extensive experiments on Microsoft COCO~\cite{chen2015microsoft} on 3 variations of image caption generation task, as in~\cite{lu2018neural} and trained semantic classifier on Visual Genome~\cite{lu2016visual}.



\textbf{COCO}~\cite{chen2015microsoft} is widely used for image captioning which contains 164062 images with 5 human-annotated captions for each. We follow standard practice to prepare the vocabulary list by collecting the words that occurred at least five times in the training set, leading to 9488 words in total.


\textbf{Visual Genome}~\cite{lu2016visual} is commonly used for modelling the interactions between objects in an image, which is composed of over 108K images and each image contains around 35 objects, 26 attributes and 21 pairwise relationships. Regarding the semantic relationship classifier, we select top 15 frequent relationship categories in training set and the classifier is thus trained over 15 relation classes with a non-relation class. 



\textbf{Evaluation metrics.} In this paper, we adopt 3 evaluation metrics, BLEU@N~\cite{papineni2002bleu}, CIDEr~\cite{vedantam2015cider} and ROUGE-L~\cite{lin2004rouge}. 


\textbf{Implementation details.} Our R-NBT is implemented with PyTorch. The dimension of region-level features and relationship features are both 1024. We follow the other parameter settings proposed in  NBT~\cite{lu2018neural}. In inference stage, we use beam search and set beam size to 3 to generate sentences. 

\subsection{Standard Image Captioning}

In standard image captioning task, we exploit the widely used split~\cite{karpathy2015deep} - 113287/5000/5000 images for train/val/test sets. 

\begin{table}
    \centering
    \caption{Comparison with State-of-the-art methods on widely used test portion of splits in~\cite{karpathy2015deep} on COCO for standard image captioning. ('Pre' shows the dataset used for pre-training Faster R-CNN, 'B@N', 'C' and 'R' represent BLEU@N, CIDEr and ROUGE-L, respectively. NBT* is our re-produced neural baby talk~\cite{lu2018neural}.)}
    \label{tab:perfomanceothermodel}
    \begin{tabular}{cccccc}
    \hline
    Method                                                  & Pre   & B@1           & B@4           & C              & R             \\ \hline
    
    NIC \cite{vinyals2015show}             & -                                                                                      & 66.6          & 20.3          & -              & -             \\
    Soft-Att \cite{xu2015show}             & -                                                                                      & 71.8          & 25.0          & -              & -             \\
    U-D$_{res}$ \cite{anderson2018bottom}  & -                                                                                     & 74.5          & 33.4          & 105.4          & 54.4          \\
    NBT$^*$ \cite{lu2018neural}                & COCO                                                                                   & 74.4          & 33.1          & 102.6          & 54.6          \\
    NTT \cite{zohourianshahzadi2020neural} & COCO                                                                                   & 73.9         & 32.9         & 101.7         & -             \\ \hline
    R-NBT                                                   & COCO                                                                                   & \textbf{75.5}          & \textbf{34.7}          & \textbf{107.0}          & \textbf{55.6}          \\ \hline
    \end{tabular}
\end{table}

\textbf{Comparison with the state-of-the-art works.} Table~\ref{tab:perfomanceothermodel} shows the results comparison with the state-of-the-art models. It indicates that R-NBT outperforms all the other methods. Although GCN-LSTM \cite{yao2018exploring} introduced in Sect.~\ref{sec:intro} achieves 77.4, 37.1, 117.1 and 57.2 on B@1, B@4, C and R,  our R-NBT and GCN-LSTM both take advantage of visual relationships in generating captions, which demonstrates the effectiveness of modelling relationships on image representations. GCN-LSTM outperforms our R-NBT slightly, as its Faster R-CNN is pre-trained on Visual Genome dataset instead of COCO, which produces richer semantic/categorical information in image features. In particular, compared to COCO with only 80 categories of objects, Visual Genome consists of 33,877 classes, over 400x more.


\textbf{Ablation studies.} Table~\ref{tab:standardcaptioning} reports the results of R-NBT using different combinations of relationships, which reveals that even with a single relationship, R-NBT outperforms NBT over all evaluation metrics. Regarding R-NBT using a single relationship, R-NBT$_{imp}$ yields better performance than R-NBT$_{spa}$ and R-NBT$_{sem}$ on both BLEU4 and CIDEr. Concerning R-NBT using two types of relationships, each combination gives superior performance than using single relationship, which further validate the effectiveness of integrating relationships into feature representations. Also, by adding implicit relationships to R-NBT$_{spa}$ (or R-NBT$_{sem}$ or R-NBT$_{spa+sem}$), R-NBT$_{imp+spa}$ (or R-NBT$_{imp+sem}$ or R-NBT) improves performance by 0.6 (or 1.5 or 0.7) on CIDEr. This points out that implicit GAT captures hidden information that semantic and spatial GATs cannot get, and generates better captions. By fusing 3 relationships, R-NBT improves the baseline network by a larger margin.



\begin{table}[t]
    \centering
    \caption{Ablation study results of visual relationships on COCO for standard image captioning. (Note: footnotes are combinations of GATs.)}
    \begin{tabular}{ccccc}
    \hline
    Method            & B@1           & B@4           & C              & R             \\ \hline
    NBT*               & 74.4          & 33.1         & 102.6          & 54.6          \\ \hline
    R-NBT$_{imp}$     & 75.0          & \textbf{33.9} & \textbf{105.1} & 54.9          \\
    R-NBT$_{spa}$     & \textbf{75.1} & 33.6          & 104.8          & 54.9          \\
    R-NBT$_{sem}$     & 74.9          & 33.7          & 104.0          & \textbf{55.0} \\ \hline
    R-NBT$_{imp+spa}$ & 75.2          & 34.1          & 105.4          & 55.2          \\
    R-NBT$_{imp+sem}$ & 75.2          & \textbf{34.2}          & 105.5          & 55.2          \\
    R-NBT$_{spa+sem}$ & \textbf{75.5} & 34.1          & \textbf{106.3}          & \textbf{55.3}  \\ \hline
    \textbf{R-NBT}    & \textbf{75.5} & \textbf{34.7} & \textbf{107.0} & \textbf{55.6} \\ \hline
    \end{tabular}
    \label{tab:standardcaptioning}
\end{table}

We also study the effects of graph attention mechanism by replacing GAT to classical GCN~\cite{yao2018exploring} in Table~\ref{tab:ablationstudy}. We show that adding graph attention mechanism leads to higher performance for all two types of relationships. In particular, the spatial and semantic relationships with graph attention are improved by 1.5 and 0.4. Note that the implicit relationships are built on graph attention mechanism, making it impossible to do such ablation experiments on R-NBT$_{imp}$.

\begin{table}[t]
    \centering
    \caption{Ablative results of group attention mechanism (GA) on CIDEr for standard image captioning.}
    \begin{tabular}{c|cccc}
    \hline
    GA & R-NBT$_{spa}$        & R-NBT$_{sem}$       & R-NBT$_{imp}$       & R-NBT            \\ \hline
    \XSolidBrush              & 103.3          & 103.6          & N/A            & N/A            \\
    \textbf{\checkmark}    & \textbf{104.8} & \textbf{104.0} & \textbf{105.1} & \textbf{107.0} \\ \hline
    \end{tabular}
    \label{tab:ablationstudy}
\end{table}

\begin{figure}[!t]
    \begin{center}
    \includegraphics[width=8.5cm]{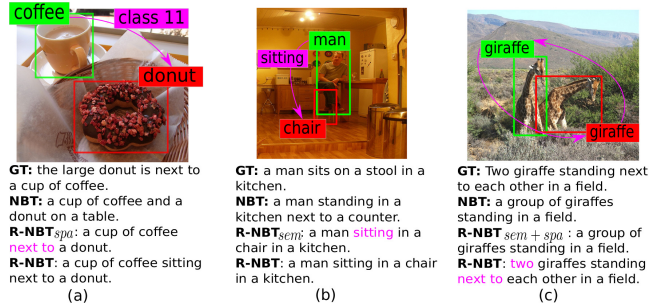}
    \end{center}
    \vspace{-0.2cm}
    \caption{Visualisations. The top row are test images with our predicted objects and relationships; The bottom row are captions of GroundTruth (GT), NBT, and generated from different variations of R-NBT.}
    \label{fig:visualization}
\end{figure}

\textbf{Visualization.} Fig.~\ref{fig:visualization} displays examples of generated captions, which shows R-NBT generates more descriptive and precise sentences in terms of relationship prediction. In Fig.~\ref{fig:visualization}, R-NBT$_{spa}$ predicts correct geometrical position ``next to" in (a) which is ignored by NBT; same to semantic relation ``sitting" for R-NBT$_{sem}$ in (b). Empirically, R-NBT$_{spa}$/R-NBT$_{sem}$ hardly model quantity of objects in an image, while R-NBT$_{imp}$ fills in the gap in (c). Clearly, ``two" giraffes is more accurate than ``a group of" as in NBT/R-NBT$_{spa+sem}$.



\subsection{Robust Image Captioning/ Novel Image Captioning}

Robust image captioning~\cite{lu2018neural} is to evaluate the performance of novel scene compositions when some relationships in the test set do not exist or only appear very few times, while in Novel image captioning~\cite{hendricks2016deep}, captions with some particular objects  are excluded. Since our model is based on the visual words to fill the slots on templates, it can seamlessly generate descriptions for images. Table~\ref{tab:robustcaptioning} and Table~\ref{tab:novelcaptioning} present results on these two tasks, which show that R-NBT performs consistently better than NBT on all metrics, and demonstrates the generalisation and robustness of our proposed methods. 


\begin{table}[!t]
    \centering
    \caption{Performance on Robust splits on COCO dataset.}
    \begin{tabular}{ccccc}
    \hline
    Method         & B@1           & B@4           & C             & R             \\ \hline
    NBT            & 73.8          & 31.6          & 93.7          & 53.1          \\ \hline
    \textbf{R-NBT} & \textbf{74.5} & \textbf{32.8} & \textbf{96.5} & \textbf{53.8} \\ \hline
    \end{tabular}
    \label{tab:robustcaptioning}
\end{table}

\begin{table}[!t]
    \centering
    \caption{Performance on Novel splits on COCO dataset.}
    \begin{tabular}{ccccc}
    \hline
    Method         & B@1           & B@4           & C             & R             \\ \hline
    NBT            & 73.5          & 31.6          & 96.5          & 53.6          \\ \hline
    \textbf{R-NBT} & \textbf{74.3} & \textbf{32.8} & \textbf{99.3} & \textbf{54.4} \\ \hline
    \end{tabular}
    \label{tab:novelcaptioning}
    \vspace*{-5mm}
\end{table}



\section{Conclusion}
\label{sec:conclusion}


This paper presents a novel relationship-based neural baby talk (R-NBT) on image caption generation tasks. R-NBT implements implicit, spatial and semantic relationships into region-level feature representations, together with graph attention network to feed into the decoder to generate the captions. Extensive experiments on COCO dataset achieve new state of the art results which validate the effectiveness of our model. For future work, we will use larger datasets with more class categories such as VG for pre-training the model.




\section{Appendix 1: More Ablation Studies.}

\textbf{Effect of parameters in Eq. \ref{eq:fusion}.} To identity the effect of the parameters $\alpha$ and $\beta$ in Eq. \ref{eq:fusion}, we evaluate R-NBT by CIDEr with different parameters in Table \ref{tab:exploringparameter}. As shown in the table, $\alpha$ and $\beta$ are set to 0.3, which the fusion weights of spatial and semantic relationships both are 30\%, the fusion weight of implicit relationships is 40\% to generate the word at step t, the performance reaches the best.

\begin{table}
\centering
\caption{The influence on the different parameters $\alpha$ and $\beta$ in fusion model with all relationships over CIDEr on COCO.}
\label{tab:exploringparameter}
\resizebox{0.5\textwidth}{!}{%
\begin{tabular}{c|c|c|c|c|c|c|c|c}
\hline
\diagbox{$\alpha$}{$\beta$} & 0.1   & 0.2   & 0.3             & 0.4   & 0.5   & 0.6   & 0.7   & 0.8   \\ \hline
0.1                         & 104.8 & 104.9 & 105.9           & 106.0 & 106.2 & 105.7 & 105.6 & 105.3 \\ \hline
0.2                         & 105.1 & 105.7 & 106.3           & 106.4 & 106.4 & 106.0 & 106.1 & -     \\ \hline
0.3                         & 105.9 & 105.6 & \textbf{107.02} & 106.8 & 106.5 & 106.2 & -     & -     \\ \hline
0.4                         & 106.1 & 106.9 & 107.0           & 106.9 & 106.7 & -     & -     & -     \\ \hline
0.5                         & 105.9 & 106.8 & 106.9           & 106.7 & -     & -     & -     & -     \\ \hline
0.6                         & 105.6 & 106.8 & 106.5           & -     & -     & -     & -     & -     \\ \hline
0.7                         & 106.0 & 106.1 & -               & -     & -     & -     & -     & -     \\ \hline
0.8                         & 105.8 & -     & -               & -     & -     & -     & -     & -     \\ \hline
\end{tabular}%
}
\end{table}

\begin{figure}[!t]

\centering
\includegraphics[scale=0.4]{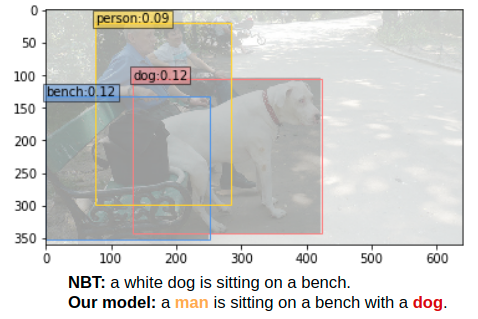}
\vspace{5pt}
\caption[Visualisation of attention maps.]{Visualisation of attention maps. We show top-3 attention weights with corresponding regions in each image. The attention weights are learned from our R-NBT model.}
\label{Fig:attentionmap}

\end{figure}

\textbf{Attention Map Visualisation.} To better understand how the graph attention mechanism helps image captioning, we visualise the attention maps learned by our R-NBT. Figure \ref{Fig:attentionmap} shows the graph attention maps on one image. Comparing the captions from NBT, it is seen that our R-NBT focuses on high weights regions to generate captions rather than treat each region equally. For instance, the caption from NBT consists of a ``dog", a ``bench" and ignore the ``person". However, our models pay more attention to three objects which are ``bench", ``dog" and ``person" to generate a caption including all three components, and the spatial relationship between person and dog is also captured. Overall, involving graph attention could help to enlarge the weights of relationships between objects when the relationships are important and contribute to a better alignment for generating captions.

\section{Appendix 2: More Visualisation}

\begin{figure*}
\centering

 
 \begin{subfigure}{0.3\textwidth}
     \centering
     \includegraphics[width=0.6\textwidth]{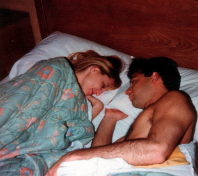}
     \caption{R-NBT: a man laying on a bed with a woman.}
 \end{subfigure}
 \hfill
 \begin{subfigure}{0.3\textwidth}
     \centering
     \includegraphics[width=0.7\textwidth]{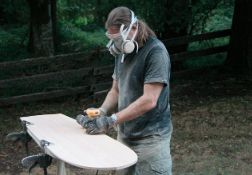}
     \caption{R-NBT: a woman standing next to a table with a surfboard.}
 \end{subfigure}
 \hfill
 \begin{subfigure}{0.3\textwidth}
     \centering
     \includegraphics[width=0.7\textwidth]{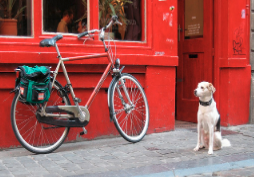}
     \caption{R-NBT: a man riding a skateboard down a street.}
 \end{subfigure}
 
 \begin{subfigure}{0.3\textwidth}
     \centering
     \includegraphics[width=0.7\textwidth]{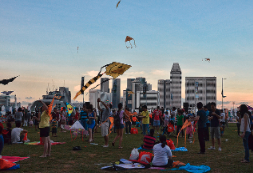}
     \caption{R-NBT: a large group of people flying kites in a field.}
 \end{subfigure}
 \hfill
 \begin{subfigure}{0.3\textwidth}
     \centering
     \includegraphics[width=0.7\textwidth]{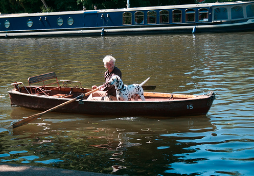}
     \caption{R-NBT: a man and a dog on a boat in the water.}
 \end{subfigure}
 \hfill
 \begin{subfigure}{0.3\textwidth}
     \centering
     \includegraphics[width=0.7\textwidth]{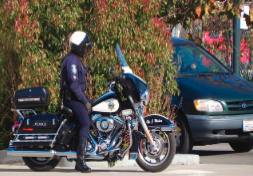}
     \caption{R-NBT: a man riding on the back of a motorcycle.}
 \end{subfigure}
 
 \begin{subfigure}{0.3\textwidth}
     \centering
     \includegraphics[width=0.7\textwidth]{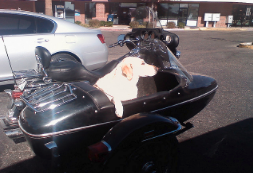}
     \caption{R-NBT: a dog sitting in the back of a motorcycle.}
 \end{subfigure}
 \hfill
 \begin{subfigure}{0.3\textwidth}
     \centering
     \includegraphics[width=0.7\textwidth]{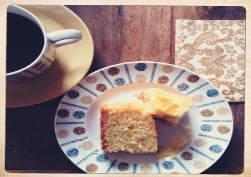}
     \caption{R-NBT: a piece of cake on a plate with a cup of coffee.}
 \end{subfigure}
 \hfill
 \begin{subfigure}{0.3\textwidth}
     \centering
     \includegraphics[width=0.7\textwidth]{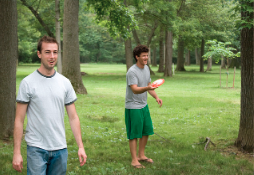}
     \caption{R-NBT: two mans playing with a frisbee in a park.}
 \end{subfigure}
 
  \begin{subfigure}{0.3\textwidth}
     \centering
     \includegraphics[width=0.7\textwidth]{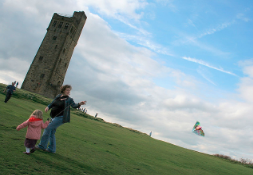}
     \caption{R-NBT: a woman and a child flying a kite in a field.}
 \end{subfigure}
 \hfill
 \begin{subfigure}{0.3\textwidth}
     \centering
     \includegraphics[width=0.7\textwidth]{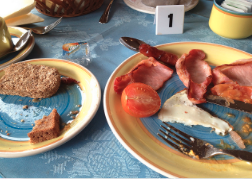}
     \caption{R-NBT: a table topped with plating of food and a fork.}
 \end{subfigure}
 \hfill
 \begin{subfigure}{0.3\textwidth}
     \centering
     \includegraphics[width=0.7\textwidth]{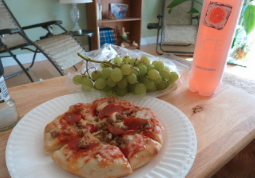}
     \caption{R-NBT: a white plate topped with a slice of pizza.}
 \end{subfigure}
 
 \begin{subfigure}{0.3\textwidth}
     \centering
     \includegraphics[width=0.7\textwidth]{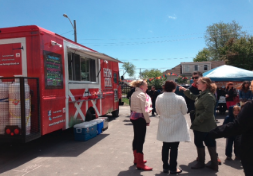}
     \caption{R-NBT: a group of people standing around a red food truck.}
 \end{subfigure}
 \hfill
 \begin{subfigure}{0.3\textwidth}
     \centering
     \includegraphics[width=0.7\textwidth]{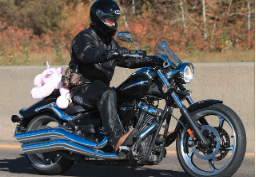}
     \caption{R-NBT: a man riding a motorcycle down a street.}
 \end{subfigure}
 \hfill
 \begin{subfigure}{0.3\textwidth}
     \centering
     \includegraphics[width=0.7\textwidth]{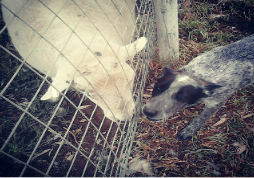}
     \caption{R-NBT: a black and white dog standing next to a white horse.}
 \end{subfigure}
 \vspace{15pt}
 \caption{More visualisation Examples from our R-NBT model.}
 \label{Fig:more_examples}
\end{figure*}

For better understanding, we show more examples of captions which are generated by our R-NBT model. See the figure \ref{Fig:more_examples}.

\bibliographystyle{IEEEbib}
\bibliography{icip21}

\end{document}